\documentclass[journal]{IEEEtran}
\usepackage{cite}
\usepackage{times}
\usepackage{amsmath,graphicx}
\usepackage{epstopdf}
\usepackage{xcolor}
\usepackage{amssymb}
\usepackage{array}
\usepackage{booktabs}
\usepackage{multirow}
\usepackage{graphicx}
\usepackage{color}
\usepackage{float}
\usepackage{stfloats}
\usepackage[misc]{ifsym}
\usepackage{stmaryrd}
\usepackage{grffile}
\usepackage[caption=false,font=footnotesize]{subfig}
\usepackage{makecell}
\usepackage{threeparttable}
\usepackage{enumitem}
\usepackage{setspace}
\usepackage{bbding}
\usepackage{caption}
\usepackage[pagebackref=true,
            breaklinks=true,
            letterpaper=true,
            bookmarks=false,
            colorlinks=true,
            linkcolor=red,
            anchorcolor=black,
            citecolor=green,
            CJKbookmarks=True]{hyperref}

\begin{document}

\markboth{Light Field Image Super-Resolution with Transformers}
{Shell \MakeLowercase{\textit{et al.}}: Bare Demo of IEEEtran.cls for IEEE Journals}
{
\title{Light Field Image Super-Resolution with Transformers}

\author{Zhengyu~Liang, Yingqian~Wang, Longguang~Wang, Jungang~Yang, Shilin~Zhou

\thanks{
This work was supported in part by the National Natural Science Foundation of China under Grants 61921001, 61972435, and 61401474.

Z.~Liang, Y.~Wang, L.~Wang, J.~Yang, and S.~Zhou are with the College of Electronic Science and Technology, National University of Defense Technology, P. R. China. Emails: \{zyliang, wangyingqian16, yangjungang\}@nudt.edu.cn. Z. Liang and Y. Wang contribute equally to this work and are co-first authors. Corresponding author: Jungang~Yang.}}

\maketitle
%%%%%%%%%%%%%%%%%%%%%%%%%%%%%%%%%%%%%%%%%%%%%%%%%%%%%%%%%%
\begin{abstract}
Light field (LF) image super-resolution (SR) aims at reconstructing high-resolution LF images from their low-resolution counterparts.
Although CNN-based methods have achieved remarkable performance in LF image SR, these methods cannot fully model the non-local properties of the 4D LF data.
In this paper, we propose a simple but effective Transformer-based method for LF image SR.
In our method, an angular Transformer is designed to incorporate complementary information among different views, and a spatial Transformer is developed to capture both local and long-range dependencies within each sub-aperture image.
With the proposed angular and spatial Transformers, the beneficial information in an LF can be fully exploited and the SR performance is boosted. We validate the effectiveness of our angular and spatial Transformers through extensive ablation studies, and compare our method to recent state-of-the-art methods on five public LF datasets. Our method achieves superior SR performance with a small model size and low computational cost. Code is available at \url{https://github.com/ZhengyuLiang24/LFT}.

\end{abstract}

%%%%%%%%%%%%%%%%%%%%%%%%%%%%%%%%%%%%%%%%%%%%%%%%%%%%%%%%%%
\section{Introduction}
\label{sec:Introduction}

\IEEEPARstart{L}{ight} field (LF) cameras record both intensity and directions of light rays, and enable many applications such as post-capture refocusing\cite{wang2018selective, jayaweera2020multi}, depth sensing \cite{wang2021enhanced, lee2018reduction}, saliency detection\cite{wang2020three} and de-occlusion \cite{DeOccNet,zhang2021removing}. Since high-resolution (HR) images are required in various applications, it is necessary to use the complementary information among different views (i.e., angular information) to achieve LF image super-resolution (SR).

\textcolor{black}{In the past few years, convolutional neural networks (CNNs) have been the dominant architectures of LF image SR.
Yoon et al. \cite{LFCNN} proposed the first CNN-based method called LFCNN to improve the resolution of LF images.
Then, some works\cite{LF-DCNN, LFNet, resLF, MegNet, LFATO} combined CNN-based method with the epipoloar geometry of LF images to improve the  SR performance.
Several works \cite{LFSSR, LF-InterNet, LF-DFnet, LF-IINet} organized LF images into the sub-aperture and macro-pixel patterns to achieve LF image SR.
Meng et al.\cite{HDDRNet} designed a high-dimensional dense residual network to learn the geometry information encoded in multiple adjacent views.}

\textcolor{black}{
Although the SR performance has been continuously improved via delicate network designs, most existing CNN-based LF image SR methods have the following two limitations.} First, these methods either use part of views to reduce the complexity of the 4D LF structure \cite{LFCNN, LF-DCNN, LFNet, resLF}, or integrate angular information without considering view position and image content \cite{LF-InterNet, LF-DFnet, LFATO}. The under-use of the rich angular information results in performance degradation especially on complex scenes (e.g., occlusions and non-Lambertain surfaces). Second, existing CNN-based methods extract spatial features by applying (cascaded) convolutions on SAIs. The local receptive field of convolutions hinders these methods to capture long-range spatial dependencies from input images. In summary, existing CNN-based LF image SR methods cannot fully exploit both angular and spatial information, and thus face a bottleneck for further performance improvement.

\textcolor{black}{Recently, Transformers have been demonstrated effective in modeling positional and long-range correlations, and were applied to various computer vision tasks such as image classification \cite{ vit}, object detection \cite{ACT, DETR}, semantic segmentation \cite{SETR}, depth estimation \cite{DPT} and super-resolution\cite{uformer, cao2021video}.}
More research and details about Transformers can be referred to Section~\ref{sec:relatedwork}.
Inspired by the recent advances of Transformers, in this paper, we propose a Transformer-based network (i.e., LFT) to address the aforementioned limitations of CNN-based methods. Specifically, we design an angular Transformer to model the relationship among different views, and design a spatial Transformer to capture both local and non-local context information within each SAI. Compared to CNN-based methods, our LFT can discriminately incorporate the information from all angular views, and capture long-range spatial dependencies in each SAI.

\textcolor{black}{The contributions of this paper can be summarized as: 1) We adapt Transformers to LF image processing and propose a simple but strong baseline for LF image SR.} 2) We propose a novel paradigm (i.e., angular and spatial Transformers) to incorporate angular and spatial information in an LF. The effectiveness of our paradigm is validated through extensive ablation studies. 3) With a small model size and low computational cost, our LFT achieves superior SR performance than other state-of-the-art methods.

\begin{figure*}
\centering
\includegraphics[width=18cm]{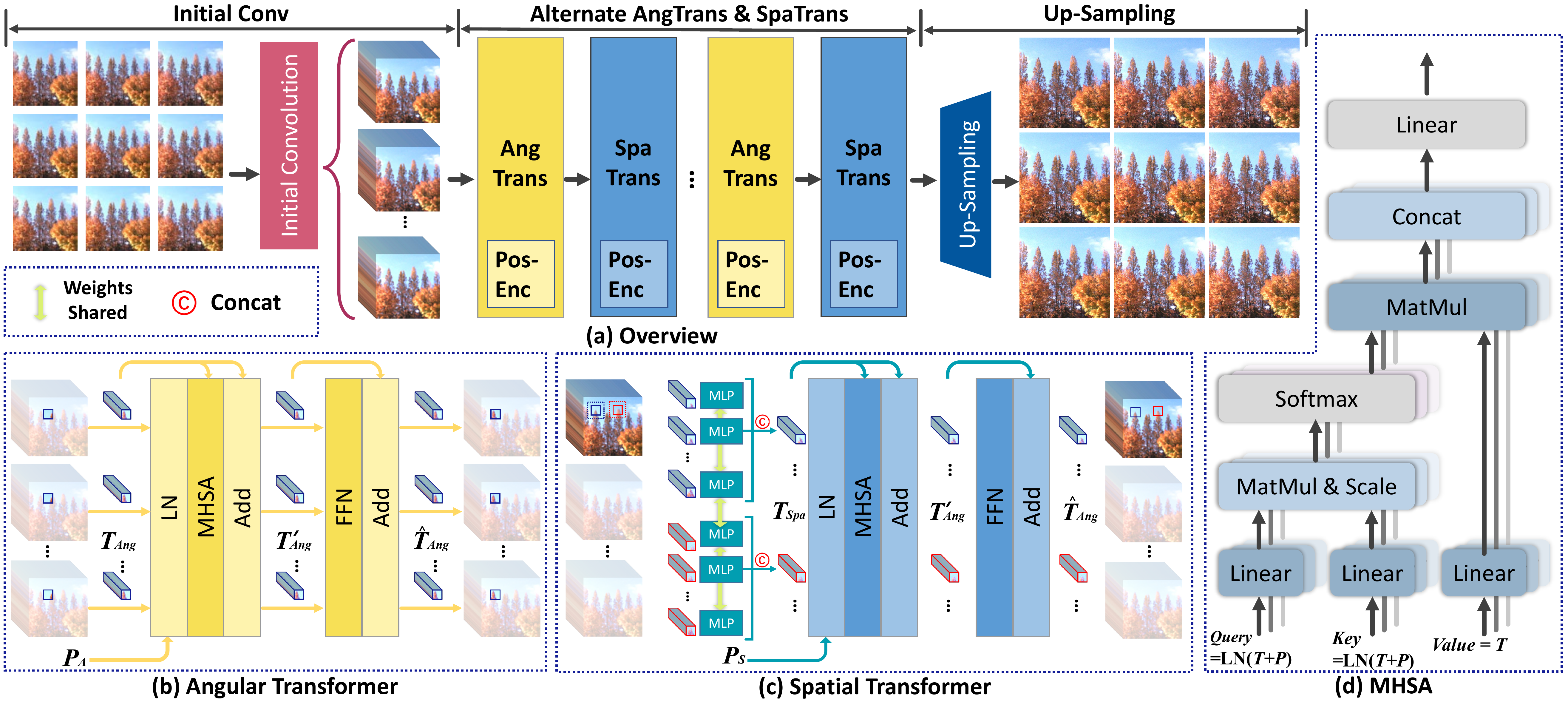}
\caption{An overview of our network.}\label{Fig_network}
%\vspace{-0.2cm}
\end{figure*}

\section{Related Work}
\label{sec:relatedwork}

\subsection{LF Image SR}

Since the CNNs are first introduced to the area of LF image SR \cite{LFCNN}, CNNs have been widely used and the performance of reconstruction has been continuously developed \cite{LF-DCNN, LFNet, resLF, MegNet, LFATO, LFSSR, LF-InterNet, LF-DFnet, LF-IINet, HDDRNet}.
Yuan et al. \cite{LF-DCNN} applied EDSR \cite{EDSR} to super-resolve each sub-aperture image (SAI) independently, and developed an EPI-enhancement network to refine the super-resolved images.
Yeung et al.\cite{LFSSR} proposed LFSSR to alternately shuffle LF features between sub-aperture and macro-pixel patterns for convolution.
Zhang et al. \cite{resLF} proposed a multi-branch residual network to incorporate the multi-directional epipolar geometry prior for LF image SR.
More recently, Zhang et al. \cite{MegNet} divided LF images into several image stacks according to multiple epipolar geometry, and fed them into different branches to effectively learn sub-pixel information.
Since both view-wise angular information and image-wise spatial information contribute to the SR performance, state-of-the-art CNN-based methods \cite{LFATO, LF-InterNet, LF-DFnet, LF-IINet, HDDRNet} designed different network structures to leverage both angular and spatial information for LF image SR.
In \cite{LF-InterNet}, the spatial and angular features can be separately extracted from LF images, and then repetitively interacted to progressively incorporate spatial and angular information.
Jin et al. \cite{LFATO} proposed an all-to-one framework to super-resolve each view using all the remaining views, and designed a structure-aware loss to preserve the parallax structure of LF images.
Meng et al. \cite{HDDRNet} designed a high-dimensional dense residual network (i.e., HDDRNet), and applied 4D convolutions to learn the geometry information encoded in multiple adjacent views.

% \vspace{-0.1cm}
\subsection{Vision Transformers}
\label{sec:Transformer}

Different from the convolutions that process an image locally, vision Transformers regards an image as a sequence of tokens, and build the relationship between the pairs of all tokens by self-attention mechanism.
Thanks to the capability of modeling long-range dependencies (i.e., the relationship of all tokens), Transformers have achieved competitive performance on many vision tasks, e.g., object detection \cite{DETR}, classification \cite{ACT} and semantic segmentation \cite{SETR}.
Among them, Carion et al. \cite{DETR} first applied Transformers to object detection to model the relations of objects and global image contexts, and directly generate the set of predictions in detection.
Dosovitskiy et al. \cite{vit} designed a pure Transformer model (i.e., ViT) for classification tasks, and ViT achieves competitive results as compared to CNN-based methods.
Zheng et al. \cite{SETR} treated semantic segmentation as a sequence-to-sequence prediction task, and designed a pure Transformer encoder and a CNN-based decoder to model global context and pixel-level segmentation, respectively.
% Chen and Yan et al.\cite{TransT} proposed a Transformer tracking method, namely TransT, which can effectively combines the template and search region features.
In the area of low-level vision, Chen et al. \cite{ipt} developed an image processing Transformer with multi-heads and multi-tails. Their method achieves state-of-the-art performance on image denoising, deraining, and SR. Wang et al.\cite{uformer} proposed a hierarchical U-shaped Transformer to capture both local and non-local context information for image restoration. Cao et al. \cite{cao2021video} proposed a Transformer-based network to exploit correlations among different frames for video SR.

\section{Method}
\label{sec:Method}

We formulate an LF as a 4D tensor $\mathcal{L} \in \mathbb{R}^{U \times V \times H \times W}$, where $U$ and $V$ represent angular dimensions, $H$ and $W$ represent spatial dimensions.
Following \cite{resLF, LFSSR, HDDRNet, LFATO, LF-InterNet}, we achieve LF image SR using SAIs distributed in a square array (i.e., $U$=$V$=$A$).
\textcolor{black}{
As shown in Fig.~\ref{Fig_network}(a), our network consists of three stages including initial feature extraction, Transformer-based feature incorporation, and up-sampling. }

% \vspace{-0.1cm}
\subsection{Angular Transformer}
\label{angTrans}
The input LF images are first processed by cascaded 3$\times$3 convolutions to generate initial features $F\in \mathbb{R}^{U \times V \times H \times W \times C}$.
The extracted features are then fed to the angular Transformer to model the angular dependencies.
\textcolor{black}{Our angular Transformer is designed to correlate highly relevant features in the angular dimension and can fully exploit the complementary information among all the input views.}

Specifically, feature $F$ is first reshaped into a sequence of angular tokens $T_{Ang} \in \mathbb{R}^{ HW \times N_A \times d_A}$, where $HW$ represents the batch dimension, $N_A$$=$$UV$ is the length of the sequence and $d_A$ denotes the embedding dimension of each angular token.
Then, we perform angular positional encoding to model the positional correlation of different views \cite{attention}, i.e.,
\begin{equation}\label{eq1}
\textcolor{black}{
P_{A}(p,2i)={\rm sin}(p/\alpha^{2i/d_A}),}
\end{equation}
\begin{equation}
\textcolor{black}{
P_{A}(p,2i\text{+}1)={\rm cos}(p/\alpha^{2i/d_A}),}
\end{equation}
where $p$$=$$\{1,2,...,A^2\}$ represents the angular position and $i$ denotes the channel index in embedding dimension.
\textcolor{black}{It is worth mentioning that each dimension of the positional encoding vector is regarded as a sinusoid, and $\alpha$ represents the wavelength of sinusoids to form a geometric progression from 2$\pi$ to $\alpha \cdot$2$\pi$ for the angular positional encoding.
Following \cite{attention}, we set $\alpha$ to 10000 in our experiments.}

\textcolor{black}{As shown in Fig.~\ref{Fig_network}(b) and Fig.~\ref{Fig_network}(d), }the angular position codes $P_{A}$ are directly added to $T_{Ang}$, and passed through a layer normalization (LN) to generate query ${Q}_A$ and key ${K}_A$, i.e., ${Q}_A$$=$${K}_A$$=$${\rm LN}(T_{Ang}+P_{A})$. Value ${V}_A$ is directly assigned as $T_{Ang}$, i.e., ${V}_A$$=$$T_{Ang}$.
\textcolor{black}{Afterwards, we apply the multi-head self-attention (MHSA) to learn the relationship among different angular tokens.
Similar to other MHSA approaches \cite{attention, ipt, vit}, the embedding dimension of ${Q}_A$, ${K}_A$ and ${V}_A$ is split into $N_H$ groups, where $N_H$ is the number of heads. }
For each attention head, the calculation can be formulated as:
\begin{equation}\label{eq3}
\begin{split}
{{H}_h} =
{\rm Softmax
  (\frac
    {\it{Q}_{\it{A,h}} {\it{W}_{{{Q},h}}}  (\it{K}_{\it{A,h}} {\it{W}_{{{K},h}}} )^{\it{T}} }
    {\sqrt{ \it{\it{d_A}/\it{N_H}} }})}
    \it{V}_{\it{A}} {\it{W}_{{\it{V},h}}},
\end{split}
\end{equation}
where $h$$=$$\{1,2,...,N_H\}$ denotes the index of head groups.
${\it{W}_{{{Q},h}}}$,  ${\it{W}_{{{K},h}}}$ and ${\it{W}_{{{V},h}}} \in \mathcal{R}^{(\it{d_A}/\it{N_H}) \times (\it{d_A}/\it{N_H})}$ are the linear projection matrices.
In summary, the MHSA can be formulated as:
\begin{equation}\label{eq4}
\begin{split}
{\rm MHSA}({Q}_A, {K}_A, {V}_A)=
[{{H}_1},...,{{H}_{N_H}} ] W_{{O}},
\end{split}
\end{equation}
where $W_{{O}} \in \mathcal{R}^{d_A \times d_A} $ is output projection matrix, $[\cdot]$ denotes the concatenation operation.

\textcolor{black}{As shown in Fig.~\ref{Fig_network}(b),} to further incorporate the correlations built by MHSA, the tokens are further fed to a feed forward network (FFN), which consists of a LN and a multi-layer perception (MLP) layer. In summary, the calculation process of our angular Transformer can be formulated as:
\begin{equation}\label{eq5}
T'_{Ang}={\rm MHSA}({Q}_A, {K}_A, {V}_A)+T_{Ang},
\end{equation}
\begin{equation}
\hat{T}_{Ang}={\rm MLP}({\rm LN}(T'_{Ang}))+T'_{Ang}.
\end{equation}

Finally, $\hat{T}_{Ang}$ is reshaped into ${F} \in \mathbb{R}^{U \times V \times H \times W \times C}$ and fed to the subsequent spatial Transformer to incorporate spatial context information.

%\vspace{-0.1cm}
\subsection{Spatial Transformer}
The goal of our spatial Transformer is to leverage both local context information and long-range spatial dependencies within each SAI. Specifically, the input feature ${F}$ is first unfolded in each 3$\times$3 neighbor region \cite{LIIF}, and then fed to an MLP to achieve local feature embedding. That is,
\begin{equation}\label{eq6}
\begin{split}
{F}'(x,y) = {\rm MLP}(\mathop{\rm concat}\limits_{x_r=\{-1,0,1\} \atop y_r=\{-1,0,1\}}{F}(x-x_r,y-y_r)),
\end{split}
\end{equation}
where $(x,y)$ denotes an arbitrary spatial coordinate on feature ${F}$.
The local assembled feature ${F}'$ is then cropped into overlapping spatial tokens $T_{Spa}\in \mathbb{R}^{ UV \times N_S \times d_S}$, where $UV$ denotes the batch dimension, $N_S$ represents the length of the sequence, and $d_S$ represents the embedding dimension of the spatial tokens.

\textcolor{black}{By performing feature unfolding and overlapped cropping, the local context information can be fully integrated into the generated spatial tokens, which enables our spatial Transformer to model both local and non-local dependencies.} To further model the spatial position information, we perform 2D positional encoding on spatial tokens:
\begin{equation}\label{eq7}
\textcolor{black}{
P_\textit{S}(p_x,p_y,2j)  = {\rm sin}({p_x}/{\alpha^{2j/d_S}})
+ {\rm sin}({p_y}/{\alpha^{2j/d_S}}),}
\end{equation}
\begin{equation}
\textcolor{black}{
P_\textit{S}(p_x,p_y,2j\text{+}1)= {\rm cos}({p_x}/{\alpha^{2j/d_S}}) + {\rm cos}({p_y}/{\alpha^{2j/d_S}}),}
\end{equation}
\textcolor{black}{
where $(p_x, p_y)$$=$$\{(1,1),...,(H,W)\}$ denotes the spatial position, $j$ denotes the index in the embedding dimension and parameter $\alpha$ is set as 10000. }Then, ${Q}_S$, ${K}_S$ and ${V}_S$ can be calculated according to:
\begin{equation}\label{eq8}
{Q}_S={K}_S={\rm LN}(T_{Spa}+P_{S}),
\end{equation}
\begin{equation}
{V}_S=T_{Spa}.
\end{equation}

\textcolor{black}{As shown in Fig.~\ref{Fig_network}(c),}
similar to the proposed angular Transformer, we use the MHSA and FFN to build our spatial Transformer. That is,
\begin{equation}
T'_{Spa} ={\rm MHSA}({Q}_S, {K}_S, {V}_S) + T_{Spa},
\end{equation}
\begin{equation}
\hat{T}_{Spa} ={\rm MLP}({\rm LN}(T'_{Spa}))+T'_{Spa}.
\end{equation}

Then, $\hat{T}_{Spa}$ is reshaped into $F\in \mathbb{R}^{U \times V \times H \times W \times C}$ and fed to the next angular Transformer.
After passing through all the angular and spatial Transformers, both angular and spatial information in an LF can be fully incorporated. Finally, we apply pixel shuffling \cite{PixelShuffle} to achieve feature up-sampling, and obtain the super-resolved LF image $\mathcal{L}_{out} \in \mathbb{R}^{U \times V \times \alpha H \times \alpha W}$.

% \vspace{-0.3cm}
\section{Experiments}
\label{sec:Experiments}

% \vspace{-0.3cm}

% In this section, we first introduce our implementation details, then compare our LFT to state-of-the-art SR methods. Finally, we conduct ablation studies to validate our design choices.

\begin{table*}[t]
%\vspace{-.05in}
\scriptsize
\centering
\renewcommand\arraystretch{1.3}
\caption{PSNR/SSIM values achieved by different methods for 2$\times$ and 4$\times$SR. The best results are in  {\textbf{bold faces}}.
% and the second best results are in \textcolor{black}{black}.
}
\label{Table_sota}
%\vspace{-0.2cm}
\begin{tabular}{l|ccccc|ccccc}
\hline
\multirow{2}*{Methods} &
\multicolumn{5}{c|}{$2\times$}&
\multicolumn{5}{c}{$4\times$}\\
\cline{2-11}
& \textit{EPFL} & \textit{HCInew} & \textit{HCIold} & \textit{INRIA} & \textit{STFgantry} & \textit{EPFL} & \textit{HCInew} & \textit{HCIold} & \textit{INRIA} & \textit{STFgantry}\\
\hline

\textit{Bicubic}    & 29.74/0.941 & 31.89/0.939 & 37.69/0.979 & 31.33/0.959 & 31.06/0.954
                    & 25.14/0.833 & 27.61/0.853 & 32.42/0.931 & 26.82/0.886 & 25.93/0.847 \\
\textit{VDSR} \cite{VDSR}
                    & 32.50/0.960 & 34.37/0.956 & 40.61/0.987 & 34.43/0,974 & 35.54/0.979
                    & 27.25/0.878 & 29.31/0.883 & 34.81/0.952 & 29.19/0.921 & 28.51/0.901\\
\textit{EDSR} \cite{EDSR}
                    & 33.09/0.963 & 34.83/0.959 & 41.01/0.988 & 34.97/0.977 & 36.29/0.982
                    & 27.84/0.886 & 29.60/0.887 & 35.18/0.954 & 29.66/0.926 & 28.70/0.908\\
\textit{RCAN} \cite{RCAN}
                    & 33.16/0.964 & 34.98/0.960 & 41.05/0.988 & 35.01/0.977 & 36.33/0.983
                    & 27.88/0.886 & 29.63/0.888 & 35.20/0.954 & 29.76/0.927 & 28.90/0.911\\
\textit{resLF}\cite{resLF}
                    & 33.62/0.971 & 36.69/0.974 & 43.42/0.993 & 35.39/0.981 & 38.36/0.990
                    & 28.27/0.904 & 30.73/0.911 & 36.71/0.968 & 30.34/0.941 & 30.19/0.937 \\

\textit{LFSSR} \cite{LFSSR}
                    & 33.68/0.974 & 36.81/0.975 & 43.81/0.994 & 35.28/0.983 & 37.95/0.990
                    & 28.27/0.908 & 30.72/0.912 & 36.70/0.969 & 30.31/0.945 & 30.15/0.939 \\

\textit{LF-ATO} \cite{LFATO}
                    & \textcolor{black}{34.27}/\textcolor{black}{0.976}
                    & \textcolor{black}{37.24}/\textcolor{black}{0.977} & \textcolor{black}{44.20}/\textcolor{black}{0.994} & \textcolor{black}{36.15}/\textcolor{black}{0.984} & {39.64}/\textcolor{black}{0.993}
                    & {28.52}/{0.912} & 30.88/0.914 & 37.00/0.970 & {30.71}/{0.949} & 30.61/0.943\\

\textit{LF-InterNet} \cite{LF-InterNet}
                    & \textcolor{black}{34.14}/\textcolor{black}{0.972} & \textcolor{black}{37.28}/\textcolor{black}{0.977} & {44.45}/\textcolor{black}{\textbf{0.995}}
                    & \textcolor{black}{35.80}/\textcolor{black}{0.985}
                    & \textcolor{black}{38.72}/\textcolor{black}{0.992}

                    & \textcolor{black}{28.67}/\textcolor{black}{0.914} & \textcolor{black}{30.98}/\textcolor{black}{0.917} & \textcolor{black}{37.11}/{0.972}
                    & \textcolor{black}{30.64}/\textcolor{black}{0.949} & \textcolor{black}{30.53}/\textcolor{black}{0.943} \\

\textit{LF-DFnet} \cite{LF-DFnet}
                    & {34.44}/\textcolor{black}{0.977}
                    & {37.44}/\textcolor{black}{\textbf{0.979}}
                    & \textcolor{black}{44.23}/\textcolor{black}{0.994}
                    & {36.36}/\textcolor{black}{0.984}
                    & \textcolor{black}{39.61}/\textcolor{black}{0.993}

                    & {28.77}/{0.917}
                    & {31.23}/{0.920}
                    & {37.32}/{0.972}
                    & {30.83}/{0.950}
                    & {31.15}/{0.949} \\

\textit{MEG-Net} \cite{MegNet}
                    & {34.30}/\textcolor{black}{0.977}
					& {37.42}/{0.978}
					& {44.08}/{0.994}
					& {36.09}/\textcolor{black}{0.985}
					& {38.77}/{0.991}

					& {28.74}/{0.916}
					& {31.10}/{0.918}
					& {37.28}/{0.972}
					& {30.66}/{0.949}
					& {30.77}/{0.945}
	\\

\textit{LF-IINet} \cite{LF-IINet}
                    & \textcolor{black}{34.68}/\textcolor{black}{0.977}
                    & \textcolor{black}{37.74}/\textcolor{black}{\textbf{0.979}}
                    & \textcolor{black}{\textbf{44.84}}/\textcolor{black}{\textbf{0.995}}
                    & \textcolor{black}{36.57}/\textcolor{black}{0.985}
                    & \textcolor{black}{39.86}/\textcolor{black}{0.994}

                    & \textcolor{black}{29.11}/\textcolor{black}{0.920}
                    & \textcolor{black}{31.36}/\textcolor{black}{0.921}
                    & \textcolor{black}{37.62}/\textcolor{black}{\textbf{0.974}}
                    & \textcolor{black}{31.08}/\textcolor{black}{\textbf{0.952}}
                    & \textcolor{black}{31.21}/\textcolor{black}{0.950}
                    \\
\textit{DPT} \cite{LF_DPT}
                    & 34.48/0.976
                    & 37.35/0.977
                    & 44.31/0.994
                    & 36.40/0.984
                    & 39.52/0.993

                    & 28.93/0.917
                    & 31.19/0.919
                    & 37.39/0.972
                    & 30.96/0.950
                    & 31.14/0.949
                    \\

\textit{LFT}(ours)
                    & \textcolor{black}{\textbf{34.80}}/\textcolor{black}{\textbf{0.978}}
					& \textcolor{black}{\textbf{37.84}}/\textcolor{black}{\textbf{0.979}}
					& \textcolor{black}{44.52}/\textcolor{black}{\textbf{0.995}}
					& \textcolor{black}{\textbf{36.59}}/\textcolor{black}{\textbf{0.986}}
					& \textcolor{black}{\textbf{40.51}}/\textcolor{black}{\textbf{0.994}}

                    & \textcolor{black}{\textbf{29.25}}/\textcolor{black}{\textbf{0.921}}
					& \textcolor{black}{\textbf{31.46}}/\textcolor{black}{\textbf{0.922}}
					& \textcolor{black}{\textbf{37.63}}/\textcolor{black}{\textbf{0.974}}
					& \textcolor{black}{\textbf{31.20}}/\textcolor{black}{\textbf{0.952}}
					& \textcolor{black}{\textbf{31.86}}/\textcolor{black}{\textbf{0.955}} \\

\hline
\end{tabular}
% \vspace{-0.2cm}
\end{table*}

% \vspace{-0.4cm}
\subsection{Implementation Details}

Following \cite{LF-DFnet}, we used 5 public LF datasets \cite{EPFL, HCInew, HCIold, INRIA, STFgantry} to validate our method. All LF images in the training and test set have an angular resolution of 5$\times$5. In the training stage, we cropped LF images into patches of 64$\times$64$/$128$\times$128 for 2$\times/$4$\times$ SR, and used the bicubic downsampling approach to generate LR patches of size 32$\times$32.

We used peak signal-to-noise ratio (PSNR) and structural similarity (SSIM) \cite{SSIM} as quantitative metrics for performance evaluation. To obtain the metric score for a dataset with $M$ scenes, we calculated the metrics on the $A$$\times$$A$ SAIs of each scene separately, and obtained the score for this dataset by averaging all the  $M$$\times$$A^2$ scores.

All experiments were implemented in Pytorch on a PC with four Nvidia GTX 1080Ti GPUs. The weights of our network were initialized using the Xavier method \cite{Xavier}, and optimized using the Adam method \cite{Adam}.
\textcolor{black}{
The batch size was set to 4$/$8 for 2$\times/$4$\times$ SR. The learning rate was set to 2$\times10^{-4}$ and halved for every 15 epochs. The training was stopped after 80 epochs.}

% \vspace{-0.2cm}
\subsection{Comparison to state-of-the-art methods}
% \vspace{-0.2cm}
\label{sec:sota}
\textcolor{black}{
We compare our LFT to several state-of-the-art methods, including 3 single image SR methods \cite{VDSR,EDSR,RCAN} and 8 LF image SR methods \cite{resLF,LFSSR,LFATO,LF-InterNet,LF-DFnet, MegNet, LF-IINet, LF_DPT}.}
We retrained all these methods on the same training datasets as our LFT.

\begin{figure*}[t]
\centering
\includegraphics[width=18cm]{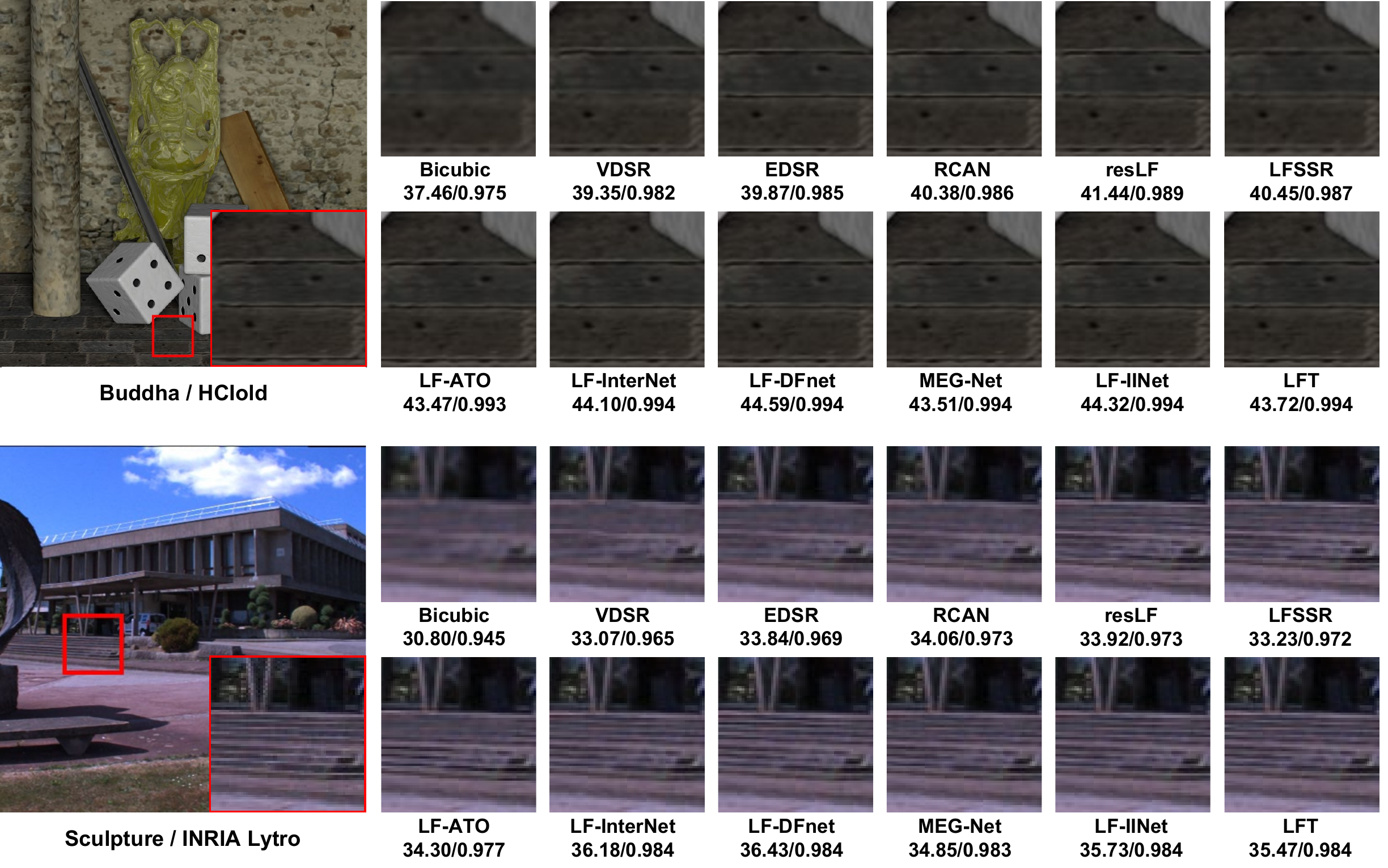}
\vspace{0.1cm}
\caption{Visual results achieved by different methods for 2$\times$SR.}\label{Fig_x2}
\end{figure*}

\begin{figure*}
\centering
\includegraphics[width=18cm]{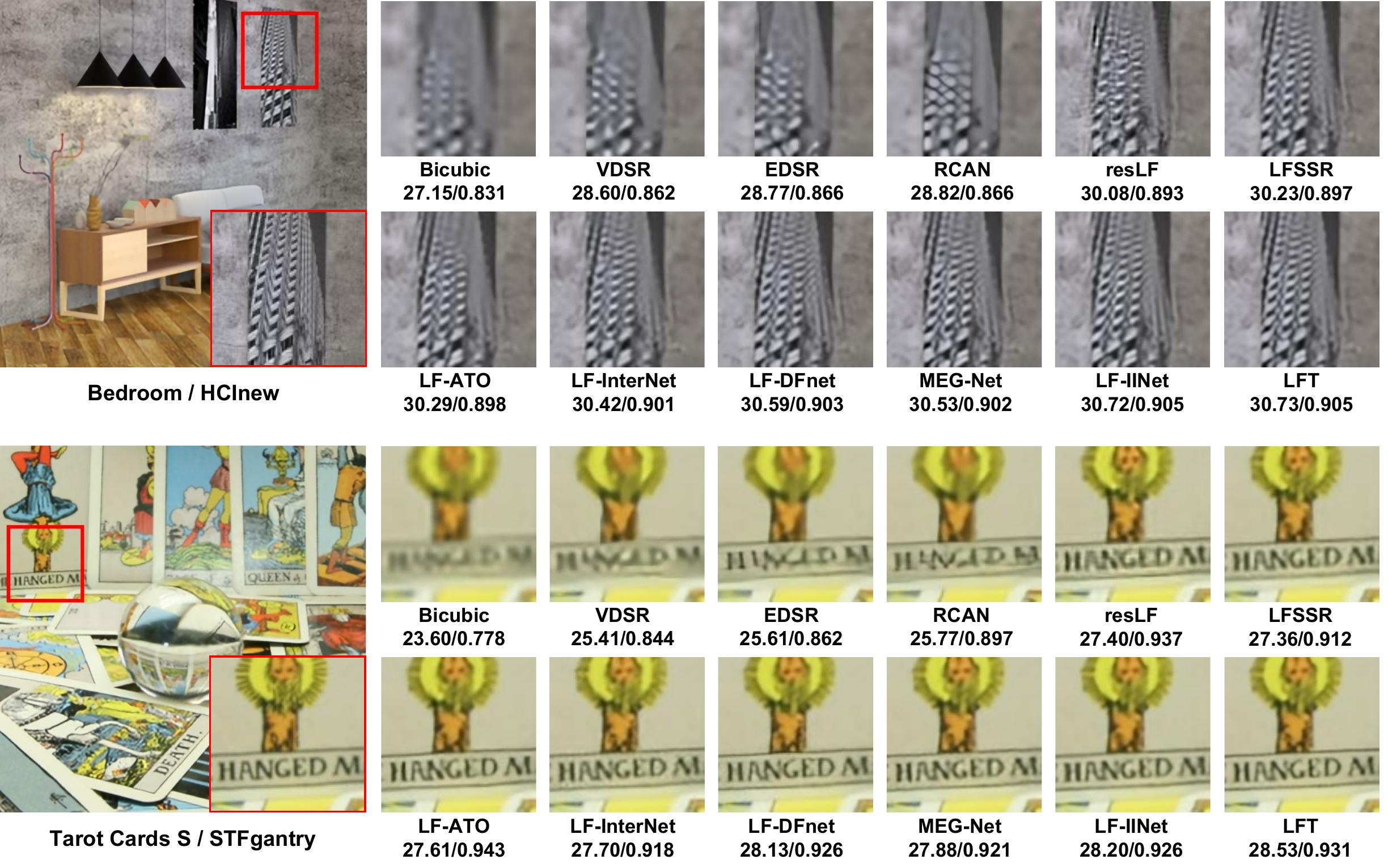}
\vspace{0.1cm}
\caption{Visual results achieved by different methods for 4$\times$SR.}\label{Fig_x4}
\end{figure*}

\subsubsection{Quantitative Results}

\textcolor{black}{Table~\ref{Table_sota} shows the quantitative results achieved by our method and other state-of-the-art SR methods.} \textcolor{black}{Our LFT achieves the competitive PSNR and SSIM results on all the 5 datasets for both 2$\times$ and 4$\times$ SR. }
Note that, the superiority of our LFT is very significant on the STFgantry dataset \cite{STFgantry} (i.e., 0.65 dB higher than the second top-performing method \cite{LF-IINet} for both 2$\times$ and 4$\times$ SR).
That is because, LF images in the STFgantry dataset have more complex structures and larger disparity variations.
By using our angular and spatial Transformers, our method can well handle these complex scenes while maintaining state-of-the-art performance on other datasets.

\subsubsection{Qualitative Results}
Figures~\ref{Fig_x2} and \ref{Fig_x4} show the qualitative results achieved by different methods.
Our LFT can well preserve the textures and details in the SR images and achieves competitive visual performance. Moreover, we provide a demo video\footnote{\href{https://wyqdatabase.s3.us-west-1.amazonaws.com/LFT_video.mp4}{https://wyqdatabase.s3.us-west-1.amazon aws.com/LFT\_video.mp4}} for a visual comparison of the angular consistency.

%\vspace{-0.0cm}
\subsubsection{Efficiency}
\textcolor{black}{
We compare our LFT to several competitive methods \cite{resLF,LFSSR,LFATO,LF-InterNet,LF-DFnet, MegNet, LF-IINet, LF_DPT} in terms of the number of parameters and FLOPs.}
\textcolor{black}{Specifically, DPT \cite{LF_DPT} is a recently proposed Transformer-based LF image SR method.}
As shown in Table~\ref{Table_Params}, compared to other methods, our LFT achieves higher accuracy with smaller model size and lower computational cost, which demonstrates the high efficiency of our method.

\begin{table}[t]
\scriptsize
\centering
\renewcommand\arraystretch{1.2}
\caption{The parameters (\#Param.), FLOPs and average PSNR/SSIM scores achieved by state-of-the-art methods for 2$\times$ and 4$\times$ SR. Note that, FLOPs is computed with an input LF of size 5$\times$5$\times$32$\times$32. The best results are in {\textbf{bold faces}}. }
% and the second best results are in \textcolor{black}{black}

% \vspace{-0.2cm}
\label{Table_Params}
\setlength{\tabcolsep}{1.2mm}
\begin{tabular}{l|ccc|ccc}
\hline
\multirow{2}*{Methods} &
\multicolumn{3}{c|}{$2\times$}&
\multicolumn{3}{c}{$4\times$}\\
\cline{2-7}
& \#Param. & FLOPs & PSNR/SSIM & \#Param. & FLOPs & PSNR/SSIM \\
\hline
\textit{resLF}
        & 7.98M & 79.63G & 37.50/0.982
        & 8.65M & 85.47G & 31.25/0.932
        \\

\textit{LFSSR}
        & \textcolor{black}{\textbf{0.89M}} & 91.06G & 37.51/0.983
        & 1.77M & 455.04G & 31.56/0.937
        \\

\textit{LF-ATO}
        & 1.22M & 1815.36G & 38.30/0.985
        & \textcolor{black}{1.36M} & 1898.91G & 31.54/0.938
        \\

\textit{LF-InterNet}
        & 4.91M & \textcolor{black}{38.97G}   & 38.08/0.985
        & 4.96M & \textcolor{black}{40.25G}   & 31.59/0.939
        \\

\textit{LF-DFnet}
        & 3.94M  & 57.22G & {38.42}/{0.985}
        & 3.99M  & 58.49G & {31.86}/{0.942}
        \\

\textit{MEG-Net}
                    & 1.69M
					& 48.40G
					& \textcolor{black}{38.12}/{0.985}
					
					& 1.77M
					& {102.2G}
					& {31.72}/{0.940}
					
	\\

\textit{LF-IINet}
                    & 4.84M
                    & 56.16G
                    & \textcolor{black}{38.74}/0.986

                    & 4.89M
                    & 57.42G
                    & \textcolor{black}{32.08}/\textcolor{black}{0.944}
                    \\

\textit{DPT}
                    & 3.73M
                    & 57.44G
                    & \textcolor{black}{38.40}/0.985

                    & 3.78M
                    & 58.64G
                    & \textcolor{black}{31.92}/\textcolor{black}{0.941}
                    \\
% \hline
\textit{LFT}(ours)
        & \textcolor{black}{1.11M} & \textcolor{black}{\textbf{28.00G}}
        & \textcolor{black}{\textbf{38.85}}/\textcolor{black}{\textbf{0.986}}
        & \textcolor{black}{\textbf{1.16M}} & \textcolor{black}{\textbf{29.45G}}
        & \textcolor{black}{\textbf{32.28}}/\textcolor{black}{\textbf{0.945}}
        \\
\hline

\end{tabular}
\end{table}

\begin{table}[t]
%\vspace{-.05in}
\scriptsize
\centering
\renewcommand\arraystretch{1.2}
\caption{PSNR results achieved on the EPFL \cite{EPFL}, HCIold \cite{HCIold} and INRIA \cite{INRIA} datasets by several variants of LFT for $4\times$SR.
Note that, AngTr and SpaTr represent models using angular Transformer and spatial Transformer, respectively. AngPos and SpaPos denote models using positional encoding in AngTr and SpaTr, respectively. \#Param. represents the number of parameters of different variants.
}
%\vspace{-0.2cm}
\label{Table_AngTrans}
\setlength{\tabcolsep}{1.5mm}
\begin{tabular}{l|cccc|c|ccc}
\hline
 & AngTr  &AngPos & SpaTr &SpaPos  & \#Param. & \textit{EPFL} & \textit{HCIold} & \textit{INRIA} \\
\hline
1 &   & &   &      & 1.49M & 28.63 & 37.00 & 30.66\\

2 & \checkmark &  &    &      & 1.42M & 28.85 & 37.29 & 30.93\\
3 & \checkmark & \checkmark   &   &      & 1.42M & \textcolor{black}{28.98} & \textcolor{black}{37.38} & \textcolor{black}{30.93}\\
4 &   &     & \checkmark &      & 1.28M & 28.93 & 37.30 & 30.97\\
5 &   &     & \checkmark & \checkmark     & 1.28M & 28.95 & 37.41 & 30.98\\
\hline
6 & \checkmark & \checkmark      & \checkmark & \checkmark    & \textcolor{black}{\textbf{1.16M}} & \textcolor{black}{\textbf{29.25}} & \textcolor{black}{\textbf{37.46}} & \textcolor{black}{\textbf{31.20}}\\

\hline
\end{tabular}
\end{table}

\begin{figure*}[t]
\centering
\includegraphics[width=14cm]{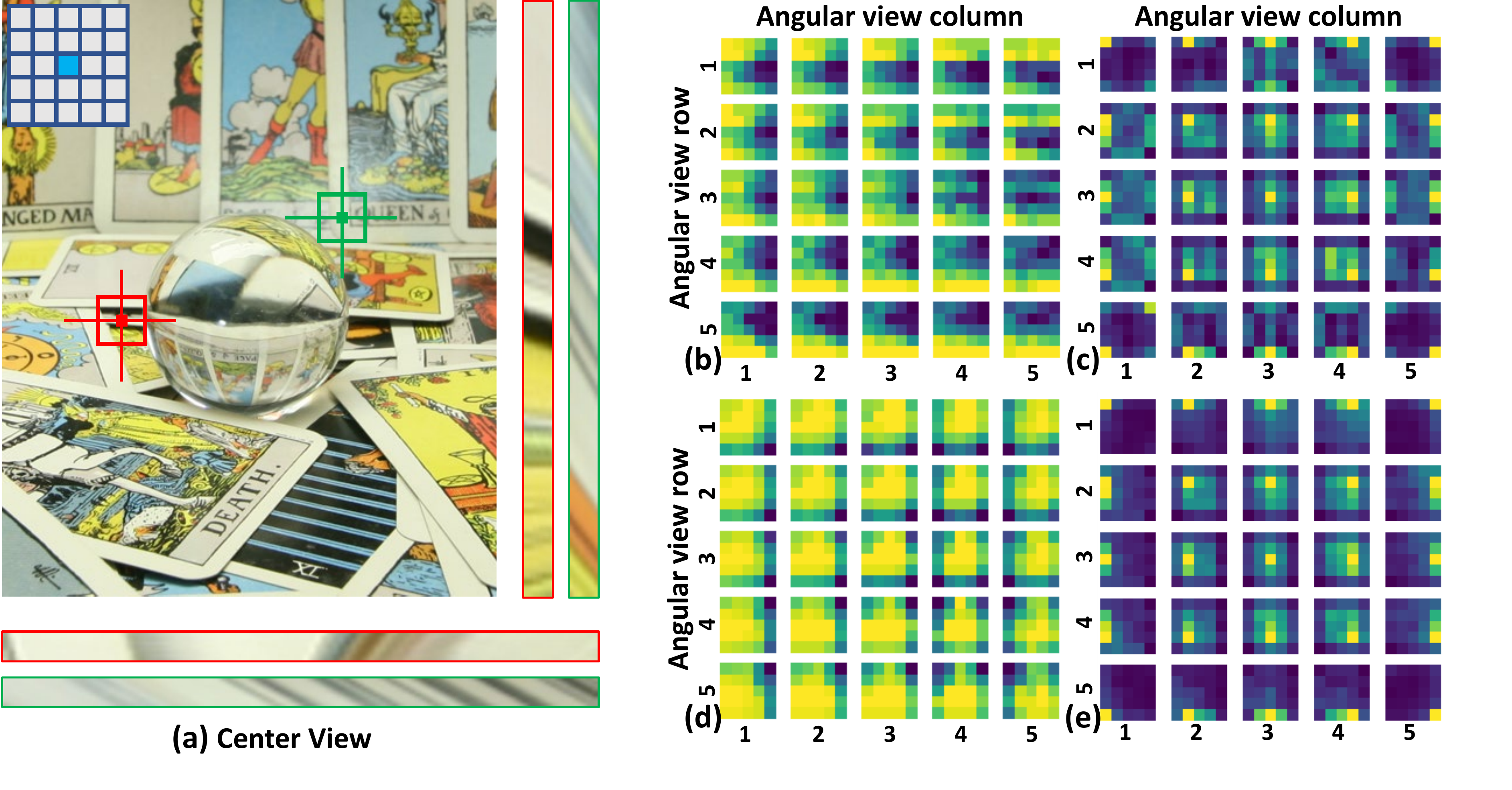}
\caption{
\textcolor{black}{
Visualization of attention maps (i.e., local angular similarities) generated by our angular Transformer and LF-InterNet\cite{LF-InterNet}.} (a) Center view and corresponding epipolar plane images of scene \textit{Cards}\cite{STFgantry}.
(b) and (c) are local angular similarity maps of the red patch in (a) by our LFT and LF-InterNet, respectively.
Similarly, (d) and (e) are of the green patch by our LFT and LF-InterNet, respectively.
Note that, each tile in the map illustrates the similarities between the current view (at the same position as the tile) and all the views (different pixels within the tile).}
\label{fig:attn_weights} %% label for entire figure
%\vspace{-0.3cm}
\end{figure*}

%%%%%%%%%%%%%%%%%%%%%%%%%%%%%%%%%%%%%%%%%%%%%%%%%%%%%%%%%%

%\vspace{-0.2cm}
\subsection{Ablation Study}

We introduce several variants with different architectures to validate the effectiveness of our method.
As shown in Table~\ref{Table_AngTrans}, we first introduce a baseline model (i.e., \textit{model-1}) without using angular and spatial Transformers,
then separately added angular Transformer (i.e., \textit{model-2}) and spatial Transformer (i.e., \textit{model-4}) to the baseline model.
Moreover, we introduce \textit{model-3} and \textit{model-5} to validate the effectiveness of angular and spatial positional encoding.

\subsubsection{Angular Transformer}
We compare the performance of \textit{model-2} to \textit{model-1} and \textit{model-6} to \textit{model-5} to validate the effectiveness of the angular Transformer. As shown in Table~\ref{Table_AngTrans}, by using the angular Transformer, \textit{model-2} achieves a 0.2$\sim$0.3 dB PSNR improvements over \textit{model-1}. When the angular positional encoding is introduced, \textit{model-3} can further achieve a 0.1dB improvement over \textit{model-2} on the EPFL \cite{EPFL} and HCIold \cite{HCIold} datasets.
\textcolor{black}{By comparing the performance of  \textit{model-5} and \textit{model-6}, we can see that removing the angular Transformer (and angular positional encoding) from our LFT will cause a notable PSNR drop (around 0.2 dB).} The above experiments demonstrate that our angular Transformer and angular positional encoding are beneficial to the SR performance.

Moreover, we investigate the spatial-aware modeling capability of our angular Transformer by visualizing the local angular attention maps.
Specifically, we selected two patches from scene \textit{Cards} \cite{STFgantry}, and obtained the attention maps (a 25$\times$25 matrix for a spatial location in a 5$\times$5 LF) produced by the MHSA in the first angular Transformer at each spatial location in the patches.
Note that, larger values in the attention maps represent higher similarities between a pair of angular tokens. We then define ``local angular attention'' by calculating the ratios of similar tokens (with attention scores larger than 0.025) in the selected patches.
\textcolor{black}{
Finally, we visualize the local angular attention map in  Fig.~\ref{fig:attn_weights} by assembling the calculated attention values according to their angular coordinates.
It can be observed in Fig.~\ref{fig:attn_weights}(b) that the attention values in the occlusion area (red patch) are distributed unevenly, where the non-occluded pixels share larger values.
It demonstrates that our angular Transformer can adapt to different image contents and achieve spatial-aware angular modeling.}

\textcolor{black}{Similarly, we calculate the self-correlation of the feature generated by the LF-InterNet\cite{LF-InterNet} (CNN-based method) in angular dimensions, and the similarity maps of red patch and green patch are shown in Fig.~\ref{fig:attn_weights}(c) and Fig.~\ref{fig:attn_weights}(d), respectively.
It can be observed that the view-wise correlation of LF-InterNet is not obvious, i.e., each view has higher similarities with its adjacent views only.
}

\subsubsection{Spatial Transformer}
We demonstrate the effectiveness of the spatial Transformer by comparing the performance of \textit{model-4} to \textit{model-1} and \textit{model-6} to \textit{model-3}. As shown in Table~\ref{Table_AngTrans}, \textit{model-4} achieves a 0.3 dB improvements in PSNR over \textit{model-1}. \textcolor{black}{Moreover, when the spatial Transformer is removed from our LFT, \textit{model-3} suffers a notable performance degradation (0.08$\sim$0.27 dB in PSNR).} That is because, compared to cascaded convolutions, the proposed spatial Transformer can better exploit long-range context information with a global receptive field, and can capture more beneficial spatial information for image SR.

\section{Conclusion}
In this paper, we propose a Transformer-based network (i.e., LFT) for LF image SR. By using our proposed angular and spatial Transformers, the complementary angular information among all the views and the long-range spatial dependencies within each SAI can be effectively incorporated. Experimental results have demonstrated the superior performance of our LFT over state-of-the-art CNN-based SR methods.

\bibliographystyle{IEEEtran}
\bibliography{LFT}

\begin{thebibliography}{10}
\providecommand{\url}[1]{#1}
\csname url@samestyle\endcsname
\providecommand{\newblock}{\relax}
\providecommand{\bibinfo}[2]{#2}
\providecommand{\BIBentrySTDinterwordspacing}{\spaceskip=0pt\relax}
\providecommand{\BIBentryALTinterwordstretchfactor}{4}
\providecommand{\BIBentryALTinterwordspacing}{\spaceskip=\fontdimen2\font plus
\BIBentryALTinterwordstretchfactor\fontdimen3\font minus
  \fontdimen4\font\relax}
\providecommand{\BIBforeignlanguage}[2]{{%
\expandafter\ifx\csname l@#1\endcsname\relax
\typeout{** WARNING: IEEEtran.bst: No hyphenation pattern has been}%
\typeout{** loaded for the language `#1'. Using the pattern for}%
\typeout{** the default language instead.}%
\else
\language=\csname l@#1\endcsname
\fi
#2}}
\providecommand{\BIBdecl}{\relax}
\BIBdecl

\bibitem{wang2018selective}
Y.~Wang, J.~Yang, Y.~Guo, C.~Xiao, and W.~An, ``Selective light field
  refocusing for camera arrays using bokeh rendering and superresolution,''
  \emph{IEEE Signal Processing Letters}, vol.~26, no.~1, pp. 204--208, 2018.

\bibitem{jayaweera2020multi}
S.~Jayaweera, C.~Edussooriya, C.~Wijenayake, P.~Agathoklis, and L.~Bruton,
  ``Multi-volumetric refocusing of light fields,'' \emph{IEEE Signal Processing
  Letters}, vol.~28, pp. 31--35, 2020.

\bibitem{wang2021enhanced}
W.~Wang, Y.~Lin, and S.~Zhang, ``Enhanced spinning parallelogram operator
  combining color constraint and histogram integration for robust light field
  depth estimation,'' \emph{IEEE Signal Processing Letters}, vol.~28, pp.
  1080--1084, 2021.

\bibitem{lee2018reduction}
J.~Lee and R.~Park, ``Reduction of aliasing artifacts by sign function
  approximation in light field depth estimation based on foreground--background
  separation,'' \emph{IEEE Signal Processing Letters}, vol.~25, no.~11, pp.
  1750--1754, 2018.

\bibitem{wang2020three}
A.~Wang, ``Three-stream cross-modal feature aggregation network for light field
  salient object detection,'' \emph{IEEE Signal Processing Letters}, vol.~28,
  pp. 46--50, 2020.

\bibitem{DeOccNet}
Y.~Wang, T.~Wu, J.~Yang, L.~Wang, W.~An, and Y.~Guo, ``Deoccnet: Learning to
  see through foreground occlusions in light fields,'' in \emph{Winter
  Conference on Applications of Computer Vision (WACV)}, 2020, pp. 118--127.

\bibitem{zhang2021removing}
S.~Zhang, Z.~Shen, and Y.~Lin, ``Removing foreground occlusions in light field
  using micro-lens dynamic filter,'' in \emph{Proceedings of the International
  Joint Conference on Artificial Intelligence (IJCAI)}, 2021, pp. 1302--1308.

\bibitem{LFCNN}
Y.~Yoon, H.~Jeon, D.~Yoo, J.~Lee, and I.~Kweon, ``Light-field image
  super-resolution using convolutional neural network,'' \emph{IEEE Signal
  Processing Letters}, vol.~24, no.~6, pp. 848--852, 2017.

\bibitem{LF-DCNN}
Y.~Yuan, Z.~Cao, and L.~Su, ``Light-field image superresolution using a
  combined deep cnn based on epi,'' \emph{IEEE Signal Processing Letters},
  vol.~25, no.~9, pp. 1359--1363, 2018.

\bibitem{LFNet}
Y.~Wang, F.~Liu, K.~Zhang, G.~Hou, Z.~Sun, and T.~Tan, ``Lfnet: A novel
  bidirectional recurrent convolutional neural network for light-field image
  super-resolution,'' \emph{IEEE Transactions on Image Processing}, vol.~27,
  no.~9, pp. 4274--4286, 2018.

\bibitem{resLF}
S.~Zhang, Y.~Lin, and H.~Sheng, ``Residual networks for light field image
  super-resolution,'' in \emph{Proceedings of the IEEE Conference on Computer
  Vision and Pattern Recognition (CVPR)}, 2019, pp. 11\,046--11\,055.

\bibitem{MegNet}
S.~Zhang, S.~Chang, and Y.~Lin, ``End-to-end light field spatial
  super-resolution network using multiple epipolar geometry,'' \emph{IEEE
  Transactions on Image Processing}, vol.~30, pp. 5956--5968, 2021.

\bibitem{LFATO}
J.~Jin, J.~Hou, J.~Chen, and S.~Kwong, ``Light field spatial super-resolution
  via deep combinatorial geometry embedding and structural consistency
  regularization,'' in \emph{Proceedings of the IEEE Conference on Computer
  Vision and Pattern Recognition (CVPR)}, 2020, pp. 2260--2269.

\bibitem{LFSSR}
H.~Yeung, J.~Hou, X.~Chen, J.~Chen, Z.~Chen, and Y.~Chung, ``Light field
  spatial super-resolution using deep efficient spatial-angular separable
  convolution,'' \emph{IEEE Transactions on Image Processing}, vol.~28, no.~5,
  pp. 2319--2330, 2018.

\bibitem{LF-InterNet}
Y.~Wang, L.~Wang, J.~Yang, W.~An, J.~Yu, and Y.~Guo, ``Spatial-angular
  interaction for light field image super-resolution,'' in \emph{European
  Conference on Computer Vision (ECCV)}.\hskip 1em plus 0.5em minus 0.4em\relax
  Springer, 2020, pp. 290--308.

\bibitem{LF-DFnet}
Y.~Wang, J.~Yang, L.~Wang, X.~Ying, T.~Wu, W.~An, and Y.~Guo, ``Light field
  image super-resolution using deformable convolution,'' \emph{IEEE
  Transactions on Image Processing}, vol.~30, pp. 1057--1071, 2020.

\bibitem{LF-IINet}
G.~Liu, H.~Yue, J.~Wu, and J.~Yang, ``Intra-inter view interaction network for
  light field image super-resolution,'' \emph{IEEE Transactions on Multimedia},
  pp. 1--1, 2021.

\bibitem{HDDRNet}
N.~Meng, H.~So, X.~Sun, and E.~Lam, ``High-dimensional dense residual
  convolutional neural network for light field reconstruction,'' \emph{IEEE
  transactions on pattern analysis and machine intelligence}, 2019.

\bibitem{vit}
A.~Dosovitskiy, L.~Beyer, A.~Kolesnikov, D.~Weissenborn, X.~Zhai,
  T.~Unterthiner, M.~Dehghani, M.~Minderer, G.~Heigold, S.~Gelly, J.~Uszkoreit,
  and N.~Houlsby, ``An image is worth 16x16 words: Transformers for image
  recognition at scale,'' in \emph{Proceedings of the International Conference
  on Learning and Representation (ICLR)}, 2021.

\bibitem{ACT}
M.~Zheng, P.~Gao, X.~Wang, H.~Li, and H.~Dong, ``End-to-end object detection
  with adaptive clustering transformer,'' \emph{arXiv preprint
  arXiv:2011.09315}, 2020.

\bibitem{DETR}
N.~Carion, F.~Massa, G.~Synnaeve, N.~Usunier, A.~Kirillov, and S.~Zagoruyko,
  ``End-to-end object detection with transformers,'' in \emph{European
  Conference on Computer Vision (ECCV)}.\hskip 1em plus 0.5em minus 0.4em\relax
  Springer, 2020, pp. 213--229.

\bibitem{SETR}
S.~Zheng, J.~Lu, H.~Zhao, X.~Zhu, Z.~Luo, Y.~Wang, Y.~Fu, J.~Feng, T.~Xiang,
  P.~Torr \emph{et~al.}, ``Rethinking semantic segmentation from a
  sequence-to-sequence perspective with transformers,'' in \emph{Proceedings of
  the IEEE Conference on Computer Vision and Pattern Recognition (CVPR)}, 2021,
  pp. 6881--6890.

\bibitem{DPT}
R.~Ranftl, A.~Bochkovskiy, and V.~Koltun, ``Vision transformers for dense
  prediction,'' in \emph{Proceedings of the IEEE/CVF International Conference
  on Computer Vision}, 2021, pp. 12\,179--12\,188.

\bibitem{uformer}
Z.~Wang, X.~Cun, J.~Bao, and J.~Liu, ``Uformer: A general u-shaped transformer
  for image restoration,'' \emph{arXiv preprint arXiv:2106.03106}, 2021.

\bibitem{cao2021video}
J.~Cao, Y.~Li, K.~Zhang, and L.~Van~Gool, ``Video super-resolution
  transformer,'' \emph{arXiv preprint arXiv:2106.06847}, 2021.

\bibitem{EDSR}
B.~Lim, S.~Son, H.~Kim, S.~Nah, and K.~Lee, ``Enhanced deep residual networks
  for single image super-resolution,'' in \emph{Proceedings of the IEEE
  Conference on Computer Vision and Pattern Recognition Workshops (CVPRW)},
  2017, pp. 136--144.

\bibitem{ipt}
H.~Chen, Y.~Wang, T.~Guo, C.~Xu, Y.~Deng, Z.~Liu, S.~Ma, C.~Xu, C.~Xu, and
  W.~Gao, ``Pre-trained image processing transformer,'' in \emph{Proceedings of
  the IEEE Conference on Computer Vision and Pattern Recognition (CVPR)}, 2021,
  pp. 12\,299--12\,310.

\bibitem{attention}
A.~Vaswani, N.~Shazeer, N.~Parmar, J.~Uszkoreit, L.~Jones, A.~Gomez,
  {\L}.~Kaiser, and I.~Polosukhin, ``Attention is all you need,'' in
  \emph{Advances in neural information processing systems}, 2017, pp.
  5998--6008.

\bibitem{LIIF}
Y.~Chen, S.~Liu, and X.~Wang, ``Learning continuous image representation with
  local implicit image function,'' in \emph{Proceedings of the IEEE Conference
  on Computer Vision and Pattern Recognition (CVPR)}, 2021, pp. 8628--8638.

\bibitem{PixelShuffle}
W.~Shi, J.~Caballero, F.~Husz{\'a}r, J.~Totz, A.~Aitken, R.~Bishop,
  D.~Rueckert, and Z.~Wang, ``Real-time single image and video super-resolution
  using an efficient sub-pixel convolutional neural network,'' in
  \emph{Proceedings of the IEEE Conference on Computer Vision and Pattern
  Recognition (CVPR)}, 2016, pp. 1874--1883.

\bibitem{VDSR}
J.~Kim, J.~Lee, and K.~Lee, ``Accurate image super-resolution using very deep
  convolutional networks,'' in \emph{Proceedings of the IEEE Conference on
  Computer Vision and Pattern Recognition (CVPR)}, 2016, pp. 1646--1654.

\bibitem{RCAN}
Y.~Zhang, K.~Li, K.~Li, L.~Wang, B.~Zhong, and Y.~Fu, ``Image super-resolution
  using very deep residual channel attention networks,'' in \emph{European
  Conference on Computer Vision (ECCV)}, 2018, pp. 286--301.

\bibitem{LF_DPT}
S.~Wang, T.~Zhou, Y.~Lu, and H.~Di, ``Detail preserving transformer for light
  field image super-resolution,'' in \emph{Proceedings of the AAAI Conference
  on Artificial Intelligence,}, 2022.

\bibitem{EPFL}
M.~Rerabek and T.~Ebrahimi, ``New light field image dataset,'' in
  \emph{International Conference on Quality of Multimedia Experience (QoMEX)},
  2016.

\bibitem{HCInew}
K.~Honauer, O.~Johannsen, D.~Kondermann, and B.~Goldluecke, ``A dataset and
  evaluation methodology for depth estimation on 4d light fields,'' in
  \emph{Asian Conference on Computer Vision (ACCV)}.\hskip 1em plus 0.5em minus
  0.4em\relax Springer, 2016, pp. 19--34.

\bibitem{HCIold}
S.~Wanner, S.~Meister, and B.~Goldluecke, ``Datasets and benchmarks for densely
  sampled 4d light fields.'' in \emph{Vision, Modelling and Visualization
  (VMV)}, vol.~13.\hskip 1em plus 0.5em minus 0.4em\relax Citeseer, 2013, pp.
  225--226.

\bibitem{INRIA}
M.~Pendu, X.~Jiang, and C.~Guillemot, ``Light field inpainting propagation via
  low rank matrix completion,'' \emph{IEEE Transactions on Image Processing},
  vol.~27, no.~4, pp. 1981--1993, 2018.

\bibitem{STFgantry}
V.~Vaish and A.~Adams, ``The (new) stanford light field archive,''
  \emph{Computer Graphics Laboratory, Stanford University}, vol.~6, no.~7,
  2008.

\bibitem{SSIM}
Z.~Wang, A.~Bovik, H.~Sheikh, and E.~Simoncelli, ``Image quality assessment:
  from error visibility to structural similarity,'' \emph{IEEE Transactions on
  Image Processing}, vol.~13, no.~4, pp. 600--612, 2004.

\bibitem{Xavier}
X.~Glorot and Y.~Bengio, ``Understanding the difficulty of training deep
  feedforward neural networks,'' in \emph{Proceedings of the International
  Conference on Artificial Intelligence and Statistics (AISTATS)}, 2010, pp.
  249--256.

\bibitem{Adam}
D.~Kingma and J.~Ba, ``Adam: A method for stochastic optimization,''
  \emph{Proceedings of the International Conference on Learning and
  Representation (ICLR)}, 2015.

\end{thebibliography}
\markboth{Light Field Image Super-Resolution with Transformers}
{Shell \MakeLowercase{\textit{et al.}}: Bare Demo of IEEEtran.cls for IEEE Journals}
}
\end{document}